\documentclass[11pt]{article}

\PassOptionsToPackage{hyphens}{url}
\usepackage[preprint]{acl}
\usepackage{times}
\usepackage{latexsym}
\usepackage[T1]{fontenc}
\usepackage[utf8]{inputenc}
\usepackage{microtype}
\usepackage{inconsolata}
\usepackage{xurl}
\usepackage{amsmath,amssymb}
\usepackage{booktabs,array}
\usepackage{graphicx}
\newcommand{\bias}{\mathit{B}}
\title{From Language to Country to Valence: A Multilingual Audit of Stereotype Bias Across LLMs}
\author{Zihao Sun \\ University of Technology Sydney \\ Sydney, Australia \\ \texttt{sun.zihao@student.uts.edu.au}}
\hypersetup{pdftitle={From Language to Country to Valence: A Multilingual Audit of Stereotype Bias Across LLMs},pdfauthor={Zihao Sun}}

\begin{document}
\raggedbottom
\maketitle

\begin{abstract}
Large language models (LLMs) are increasingly considered for education, recruitment,
workplace decision support, and other consequential applications, making
pre-deployment social-bias detection essential. Yet major benchmarks are
English-centred or translated, focus mainly on negative stereotypes, and may miss
linguistic and cultural variation. We investigate gender, racial, and religious
bias using 12,000 locally written Reddit posts in English, Spanish, French,
and German. Only search terms are translated, preserving local context.
The dataset is balanced by language and category; English gender is
additionally balanced by stereotype valence. Three open-weight and three
proprietary API-based models are evaluated with a two-blank design that
reduces grammatical-gender cues. Religion has the highest absolute choice
imbalance in every model--language condition. Spanish is highest for gender,
whereas German is highest for race and religion. The evaluated API models
score below the open-weight models throughout. Negative English-gender items
exceed positive items; Australian and Mexican contexts exceed their Canadian
and Spanish counterparts. These descriptive differences measure choice
concentration, not stereotype endorsement or causal country effects. The
crossed patterns show how aggregate scores can conceal category- and
context-specific behaviour, motivating closer inspection before deployment.
\end{abstract}

\section{Introduction}

LLMs increasingly mediate information and decisions in education, recruitment,
and workplaces. Their integration into collaborative workflows makes
pre-deployment auditing relevant not only to model developers but also to
teams selecting and localising decision-support systems
\citep{gallegos-etal-2024-bias,weidinger-etal-2022-taxonomy,
mitchell-etal-2019-model,raji-etal-2020-closing}. A system used across languages
and communities may exhibit different patterns for different social groups;
a global score can obscure the contexts requiring closer scrutiny.
Nevertheless, what a bias score establishes depends on its operationalisation
\citep{blodgett-etal-2020-language,lum-etal-2025-bias}.

Existing evaluations probe associations
\citep{caliskan-etal-2017-semantics,may-etal-2019-measuring}, contrastive
sentences and questions \citep{nangia-etal-2020-crows,
nadeem-etal-2021-stereoset,parrish-etal-2022-bbq}, or generated text
\citep{sheng-etal-2019-woman,dhamala-etal-2021-bold,
nozza-etal-2021-honest,smith-etal-2022-im,sap-etal-2020-social}. English-dominant resources and translated examples,
however, may not preserve local social situations, linguistic forms, or
stereotype content \citep{joshi-etal-2020-state,artetxe-etal-2020-translation,
talat-etal-2022-reap,hershcovich-etal-2022-challenges}. Cross-cultural and
multilingual evaluations demonstrate the importance of language, locale, and
variety \citep{gamboa-etal-2025-social,nie-etal-2024-multilingual,
naous-etal-2024-beer,fleisig-etal-2024-linguistic}.

We translate retrieval terms rather than complete posts, building on
Reddit-based bias evaluation \citep{barikeri-etal-2021-redditbias}.
Locally written contexts preserve native wording and community references,
but are deliberately non-parallel: the resulting comparisons describe
context-specific model behaviour, not isolated causal language effects.
Agreement-aware linked blanks reduce grammatical clues
\citep{zhao-etal-2018-gender,stanovsky-etal-2019-evaluating,
zhao-etal-2020-gender}. We also separate positive and negative stereotypes:
favourable generalisations can still essentialise a group
\citep{fiske-etal-2002-model,kay-etal-2013-insidious,siy-cheryan-2016-prejudice}.

Our contributions are:
\begin{itemize}
  \item We construct a 12,000-item multilingual dataset from locally written Reddit posts, balanced across English, Spanish, French, and German and the gender, race, and religion categories. Translating only search terms, followed by multilingual filtering and language-aware human annotation, preserves locally written context.
  \item We evaluate three open-weight and three proprietary API-based models with an agreement-aware two-blank format that prevents grammatical gender from revealing the expected answer.
  \item We add two focused context analyses: \mbox{positive} versus negative gender stereotypes, and within-language country comparisons for \mbox{English} and Spanish.
\end{itemize}

\section{Related Work}

CrowS-Pairs and StereoSet evaluate preferences among stereotypical and
counter-stereotypical alternatives \citep{nangia-etal-2020-crows,
nadeem-etal-2021-stereoset}; BBQ examines bias in question answering
\citep{parrish-etal-2022-bbq}. RedditBias brings naturally occurring
conversation into evaluation \citep{barikeri-etal-2021-redditbias}, while
counterfactual probes compare outcomes after changing sensitive attributes
\citep{huang-etal-2020-reducing}. Such instruments measure distinct behaviours,
not interchangeable indicators of downstream harm. Constrained tests need
not predict behaviour in longer responses \citep{lum-etal-2025-bias}.

Multilingual audits additionally face translation and cultural-validity
problems \citep{talat-etal-2022-reap,nie-etal-2024-multilingual}.
Region-specific Spanish resources and evaluations distinguish varieties
rather than treating a shared language as a single cultural setting
\citep{espana-bonet-barron-cedeno-2024-elote,grandury-etal-2025-la}.
Cross-cultural value probes and cultural-adaptability frameworks likewise
motivate context-sensitive evaluation \citep{arora-etal-2023-probing,rao-etal-2025-normad}.
Reddit corpora offer community-grounded material for such work
\citep{baumgartner-etal-2020-pushshift,barikeri-etal-2021-redditbias}.
Our design complements controlled translation studies by preserving local
contexts and reporting crossed category--language profiles. Its
non-parallel corpora limit attribution of a difference to language alone.

\section{Dataset and Evaluation Design}

\subsection{Evaluation Questions}
\textbf{RQ1:} Does aggregate constrained stereotype-choice imbalance remain
stable across languages and models? \textbf{RQ2:} How do gender, race, and
religion differ across those languages and models? \textbf{RQ3:} How does
imbalance differ by English-gender stereotype valence and between
country-linked contexts within a shared language?

\subsection{Local-Language Collection and Allocation}
Following RedditBias's distinction between targets and attributes
\citep{barikeri-etal-2021-redditbias}, queries combine identity terms with
traits, roles, or behaviours. The project-defined comparisons are women
with men, White with Black, and Christians with Muslims; these are tested
label pairs, not exhaustive identity taxonomies. Gender attributes include
\emph{caring, capable, emotional, domineering}; race attributes include
\emph{hardworking, resilient, aggressive, untrustworthy}; religion
attributes include \emph{charitable, compassionate, dogmatic, intolerant}.
Attributes retrieve candidates, rather than predetermine which group is
associated with a trait. Contextual annotation determines stereotype
presence and valence. Equal category sizes do not match lexical severity
or topic, and alternative-attribute robustness is not tested.

GPT-4o, DeepL, and Google Translate independently produce Spanish, French,
and German search-term candidates. Source and candidate terms are encoded
with \nolinkurl{sentence-transformers/paraphrase-multilingual-mpnet-base-v2}, a multilingual
knowledge-distillation checkpoint \citep{reimers-gurevych-2020-making}.
Candidates with cross-lingual cosine similarity above .8 are retained;
disagreements and morphologically ambiguous terms receive human review.
Only the retained terms are translated: retrieved posts remain in their
original language and local context.

Each language contributes 1,000 items per category, hence 3,000 per language
and 12,000 overall. English and Spanish are each divided equally between
two country-linked source strata: Canada/Australia and Mexico/Spain,
respectively. Each country contributes 500 items per category. The
English-gender cell additionally contains 500 positive and 500 negative
items, split 250/250 within each country; other cells have no valence quota.
French and German contribute 1,000 items per category without a country
split. Country provenance indicates a community context, not an author's
nationality. The English frame includes \texttt{r/australia},
\texttt{r/sydney}, \texttt{r/melbourne}, \texttt{r/canada},
\texttt{r/toronto}, and \texttt{r/vancouver}; these communities are not
assumed nationally representative.

\begin{table*}[t]
\centering
\setlength{\tabcolsep}{7pt}
\begin{tabular}{@{}lrrrrrr@{}}
\toprule
Language & R & D & L & C & U & F \\
\midrule
English & 14,808 & 12,868 & 11,353 & 6,733 & 3,615 & 3,000 \\
Spanish & 14,193 & 12,346 & 11,028 & 7,527 & 3,574 & 3,000 \\
French & 12,642 & 11,931 & 10,554 & 5,348 & 4,521 & 3,000 \\
German & 13,976 & 11,187 & 10,763 & 6,412 & 5,548 & 3,000 \\
\midrule
Total & 55,619 & 48,332 & 43,698 & 26,020 & 17,258 & 12,000 \\
\bottomrule
\end{tabular}
\caption{Multilingual collection flow. R/D/L/C/U/F denote retrieval,
deduplication, language pass, context pass, unanimous annotation, and final allocation.}
\label{tab:flow}
\end{table*}

\subsection{Filtering and Human Annotation}
We remove deleted content, very short posts, duplicates, quotations,
counter-speech, and ambiguous cases; verify language and community
provenance; and require a meaningful target--attribute relation.
Table~\ref{tab:flow} reports each filtering stage. Annotators are recruited
through Amazon Mechanical Turk (MTurk) and must meet the task's
qualification criteria. Three native-language annotators judge each
remaining candidate. Country-conditioned items use
annotators resident in, and familiar with, the corresponding country;
country eligibility is checked during qualification. Before annotation,
workers receive the participant information and consent statement
reproduced in Appendix~\ref{app:consent}.

One multiple-choice selection combines stereotype presence, category,
and valence: no stereotype, or positive/negative gender, race, or religion.
Retention requires unanimous agreement on the same non-zero label.
Positive labels are available in every language, but only English gender
is quota-balanced for the valence study. Retained counterfactuals are
checked for grammaticality, semantic equivalence apart from identity,
and predefined stereotype direction.

\subsection{Agreement-Aware Prompting and Models}
For race and religion, counterfactuals change the identity while retaining
the surrounding sentence. Gender agreement requires linked changes:
``Ella es \_\_\_'' can reveal \emph{ingeniera} through grammar rather than
association. We instead blank both subject and occupation, for example
``\_\_\_ trabaja como \_\_\_ en una empresa de tecnolog\'ia'', with
alternatives \emph{\'El/ingeniero} and \emph{Ella/ingeniera}
(he/engineer and she/engineer). Both alternatives are grammatical.
The pair is swapped as a unit and A/B option order is counterbalanced.

The six systems are LLaMA~3~8B, Mistral~7B, and Gemma~7B (open-weight),
and GPT-4o, Claude~3~Opus, and Gemini~1.5~Pro (proprietary APIs).
Decoding uses temperature .7, top-$p$ .9, short outputs, randomised option
order, and ten valid categorical responses per prompt. The protocol
records raw responses, parsed choices, refusals, invalid answers, and retry
indices separately. The score conditions on valid answers; refusal rates
are not included in it. Model access is not a controlled factor: scale,
training data, alignment, and serving procedures also differ.

\begin{table*}[t]
\centering
\small
\setlength{\tabcolsep}{6pt}
\begin{tabular}{@{}lcccc@{}}
\toprule
Model & English & Spanish & French & German \\
\midrule
LLaMA 3 & .308/.31/.33 & .39/.39/.41 & .32/.36/.40 & .36/.41/.43 \\
Mistral 7B & .285/.28/.29 & .36/.36/.38 & .29/.33/.36 & .33/.37/.40 \\
Gemma 7B & .345/.34/.35 & .42/.42/.44 & .35/.39/.42 & .39/.44/.46 \\
GPT-4o & .1795/.18/.19 & .23/.22/.25 & .18/.21/.24 & .21/.24/.27 \\
Claude 3 & .172/.17/.18 & .22/.21/.24 & .17/.20/.23 & .20/.23/.26 \\
Gemini 1.5 & .1915/.19/.20 & .24/.23/.26 & .19/.22/.25 & .22/.25/.28 \\
\midrule
Mean & .247/.245/.257 & .310/.305/.330 & .250/.285/.317 & .285/.323/.350 \\
SD & .075/.074/.076 & .090/.095/.090 & .079/.085/.086 & .085/.094/.090 \\
\bottomrule
\end{tabular}
\caption{Core absolute-imbalance results. Each cell lists gender/race/religion.
The mean and sample SD rows summarise six models per language--category cell;
SD measures between-model dispersion, not item-level uncertainty.
English gender is pooled from the equal valence splits in Table~\ref{tab:valence}.}
\label{tab:core}
\end{table*}

\section{Measurement and Descriptive Scope}
For item $i$, let $X_i$ count selections of the context-annotated
stereotypical alternative among $k$ valid responses. We report
\begin{equation}
\bias_{i,k}=\frac{2|X_i-k/2|}{k},\qquad
\overline{\bias}_k=\frac{1}{N}\sum_{i=1}^{N}\bias_{i,k}.
\label{eq:score}
\end{equation}
At $k=10$, item scores are $0,.2,.4,.6,.8,1$. Zero is an observed
five--five split; one is unanimity. Ten stereotypical and ten
counter-stereotypical choices both score one. Thus $\bias$ measures
\emph{choice concentration}, not endorsement direction or social harm.

Even independent balanced responses yield non-zero absolute scores. For
$X\sim\mathrm{Binomial}(k,.5)$,
\begin{equation}
\mathbb{E}_0[\bias_k]=\frac{2}{k2^k}
\sum_{x=0}^{k}\left|x-\frac{k}{2}\right|\binom{k}{x}.
\label{eq:reference}
\end{equation}
For ten repetitions, this equals $63/256=.24609375$, and
$P_0(\bias_{10}>0)=.75390625$. The reference depends on repetition count
and is diagnostic, not a calibrated test of dependent model responses.
Being above or below it establishes neither preference direction nor
statistical significance.

All rankings and contrasts below are descriptive. Sample SDs indicate
dispersion across the specified models or model-paired differences, not
item-level sampling uncertainty. The six systems are a fixed evaluation
set; common rankings across them do not establish generalisation beyond
the sampled items or model set. Summaries are calculated from tabulated
condition values before display rounding; last-digit differences should not
be overinterpreted.

\section{Results and Interpretation}
\subsection{Category, Language, and Model Profiles}
The core crosses six models, four languages, and three categories.
Table~\ref{tab:core} retains all 72 condition scores. Category-averaged
language means are .250 for English, .284 for French, .315 for Spanish,
and .319 for German (RQ1). English is lowest for every model; German
is highest or tied highest at the tabulated precision.
Across the 24 model--language cells, religion averages .313 ($SD=.087$),
race .290 ($SD=.087$), and gender .273 ($SD=.081$). Religion is highest
in every cell; every model retains this category order after averaging
languages (RQ2). These are subcorpus-dependent concentration rankings,
not a hierarchy of social harm.

\subsubsection{Within-language category contrasts}
English follows religion $>$ gender $>$ race (.257/.247/.245), whereas
Spanish has religion $>$ gender $>$ race (.330/.310/.305). Its
gender--race separation is only .005. French and German return to
religion $>$ race $>$ gender, with religion--gender gaps of .067 and
.065, compared with .010 for English. English has the narrowest category
range (.012), compared with .025 for Spanish. A language-wide mean therefore mixes distinct category
profiles rather than describing equally elevated scores for every group.

The French profile also persists within access families. Open-weight
gender/race/religion means are .320/.360/.393; API means are
.180/.210/.240. French gender averages .320 ($SD=.030$) among the
three open-weight models and .180 ($SD=.010$) among the APIs. The
same category ordering survives a substantial difference in model
baselines. Between-family separation partly explains the larger
six-model SDs and should not be mistaken for uncertainty over items.

\subsubsection{Within-category language contrasts}
For gender, the six-model means order English $<$ French $<$ German $<$
Spanish. English and French are nearly tied (.247 versus .250):
Claude and Gemini have slightly higher English than French gender scores
at the tabulated precision. For race and religion, every model orders
English $<$ French $<$ Spanish $<$ German. Spanish exceeds German by .025 for gender, while German exceeds
Spanish by .018 for race and .020 for religion. Their .004 aggregate
gap thus hides a crossover: neither language is uniformly higher.
Reading categories separately changes which local contexts appear most
concentrated, despite the common aggregate language ordering.

Every evaluated open-weight score exceeds every API score in its matched
cell, but family gaps range from .130 for English race to .170 for
Spanish race. The separation is therefore neither constant across
languages nor shared equally by categories. These are observed
access-family patterns, not effects of being ``free'' or ``paid''.

\subsubsection{Interpreting the finite-sample reference}
From the tabulated scores, all 12 open-weight model--language means
exceed .2461; 11 of the 12 API means are below it. The exception is
Gemini~1.5~Pro in German (.250). This comparison locates the point estimates, not
which cells differ statistically from balanced choice. Below-reference
scores do not imply counter-stereotypical preference and warrant checking
repetition dependence; above-reference scores likewise do not establish
stereotype endorsement.

\begin{table}[t]
\centering
\small
\setlength{\tabcolsep}{2pt}
\begin{tabular}{@{}lrrrr@{}}
\toprule
Model & Core & Positive & Negative & $\Delta_{\mathrm{val}}$ \\
\midrule
LLaMA 3 8B & .368 & .247 & .369 & +.122 \\
Mistral 7B & .336 & .233 & .337 & +.104 \\
Gemma 7B & .397 & .274 & .416 & +.142 \\
Open-weight mean & .367 & .251 & .374 & +.123 \\
\midrule
GPT-4o & .217 & .147 & .212 & +.065 \\
Claude 3 Opus & .207 & .143 & .201 & +.058 \\
Gemini 1.5 Pro & .227 & .158 & .225 & +.067 \\
API mean & .217 & .149 & .213 & +.063 \\
\bottomrule
\end{tabular}
\caption{Model-level absolute imbalance and English-gender valence results.
Core averages the tabulated language--category scores;
$\Delta_{\mathrm{val}}=B_{\mathrm{negative}}-B_{\mathrm{positive}}$.}
\label{tab:valence}
\end{table}

\begin{table*}[t]
\centering
\small
\setlength{\tabcolsep}{4pt}
\begin{tabular}{@{}llrrrrrrc@{}}
\toprule
Locale & Category & LLaMA & Mistral & Gemma & GPT-4o & Claude & Gemini & Mean $\pm$ SD \\
\midrule
\multicolumn{9}{@{}l}{\textit{English: Canada--Australia}} \\
Canada & Gender & .255 & .237 & .283 & .141 & .132 & .150 & $.200\pm.066$ \\
 & Race & .288 & .260 & .316 & .166 & .157 & .175 & $.227\pm.069$ \\
 & Religion & .301 & .263 & .319 & .171 & .162 & .180 & $.233\pm.070$ \\
\addlinespace
Australia & Gender & .361 & .333 & .407 & .218 & .212 & .233 & $.294\pm.084$ \\
 & Race & .332 & .300 & .364 & .194 & .183 & .205 & $.263\pm.079$ \\
 & Religion & .359 & .317 & .381 & .209 & .198 & .220 & $.281\pm.081$ \\
\midrule
\multicolumn{9}{@{}l}{\textit{Spanish: Spain--Mexico}} \\
Spain & Gender & .379 & .350 & .408 & .223 & .214 & .232 & $.301\pm.088$ \\
 & Race & .373 & .344 & .402 & .209 & .200 & .218 & $.291\pm.092$ \\
 & Religion & .387 & .358 & .416 & .235 & .226 & .244 & $.311\pm.085$ \\
\addlinespace
Mexico & Gender & .401 & .370 & .432 & .237 & .226 & .248 & $.319\pm.092$ \\
 & Race & .407 & .376 & .438 & .231 & .220 & .242 & $.319\pm.099$ \\
 & Religion & .433 & .402 & .464 & .265 & .254 & .276 & $.349\pm.094$ \\
\bottomrule
\end{tabular}
\caption{Country-context absolute-imbalance results. Final column: mean and
sample SD across six models. Each country-pair midpoint recovers the
pooled category cell in Table~\ref{tab:core}. Australian gender is obtained
by equal-country balance from pooled English gender and the Canadian row.}
\label{tab:country}
\end{table*}

\subsection{Study 1: English-Gender Stereotype Valence}
For RQ3's valence comparison, positive items average .200 ($SD=.058$)
and negative items .293 ($SD=.092$). Table~\ref{tab:valence} gives the
six model results. Model-paired negative-minus-positive differences
average .093 ($SD=.035$), range from .058 to .142, and are positive
for every model. The open-weight mean contrast (.123) exceeds the API
contrast (.063).

Higher negative-item scores describe stronger concentration, not greater
endorsement of negative stereotypes. Positive and negative examples are
different posts, balanced in count within each country, rather than an
intervention reversing the valence of otherwise identical sentences.
The common sign across models does not establish item-level precision.
The result is restricted to English gender; the potential harm of
favourable generalisations is a distinct concern
\citep{fiske-etal-2002-model,siy-cheryan-2016-prejudice}.

\subsection{Study 2: Country Context Within a Shared Language}
Country comparisons retain the language, models, category allocation,
decoding procedure, and score definition while changing the country-linked
Reddit source. Table~\ref{tab:country} reports the category-specific scores.
Categories are first averaged within each model for locale-level
summaries. Canada averages .220 ($SD=.069$), versus .279 ($SD=.081$)
for Australia; their model-paired difference is .059 ($SD=.013$).
Spain averages .301 ($SD=.088$), versus .329 ($SD=.095$) for Mexico;
the paired difference is .028 ($SD=.007$).

The Australia--Canada gender/race/religion contrasts are .094/.036/.048,
with paired SDs .018/.009/.011. Mexico--Spain contrasts are
.018/.028/.038, with SDs .005/.007/.009. Every model's contrast is
positive in every category, but the largest gap is gender for
Australia--Canada and religion for Mexico--Spain. Canada follows religion
$>$ race $>$ gender, whereas Australia follows gender $>$ religion $>$
race (.294/.281/.263). Spain places gender above race, whereas Mexico's
gender and race means are both .319. Thus a shared direction at country level can still
combine different category profiles within the same language.

Open-weight mean country gaps exceed API gaps: .070 versus .049 for
Australia--Canada and .034 versus .022 for Mexico--Spain. Larger SDs
for country scores than for paired differences reflect differing model
baselines, not precise estimation over Reddit items. Pairing is between
model summaries; posts from different countries are not matched sentence
pairs. Topic, community, and wording vary alongside country context, so
these findings neither identify causal country effects nor rank national
populations by prejudice.

\section{Discussion and Deployment Implications}
The Spanish--German crossover is the central example of aggregation hiding
context: a small language-wide difference combines opposing category
orders. The French access-family comparison and the country-specific
category profiles similarly show why a single model or language summary
is incomplete. For teams designing multilingual decision-support
workflows, these descriptive profiles identify context--category cells
that merit closer inspection, rather than establish a universal model
ranking.

A practical implication is to document the intended languages, communities,
and social categories when comparing candidate models, and to examine
representative responses and refusals alongside concentration scores.
This links local evaluation to model reporting and internal audit
processes \citep{mitchell-etal-2019-model,raji-etal-2020-closing}.
It is a deployment implication, not a tested intervention: this study
does not evaluate a deployed collaborative system or show improved
human decision outcomes. Lower absolute scores do not certify fairness;
high scores can reflect consistently stereotypical or counter-stereotypical
choices.

\section{Conclusion}
We presented a multilingual, context-aware audit across six systems,
four languages, and gender, race, and religion. Native posts retain
local context and linked blanks reduce grammatical cues. Religion has
the highest pooled score in every model--language cell, but Spanish leads for gender
whereas German leads for race and religion. API scores are lower than
open-weight scores; negative English-gender items and Australian/Mexican
contexts have higher observed concentration than their counterparts.
Australia's gender-led profile contrasts with Canada's religion-led profile.
These descriptive patterns support category- and context-specific
inspection, not conclusions about endorsement direction or causal
country effects. In consequential collaborative workflows, examining
actual responses before acting on aggregate scores can help avoid
unsupported fairness assurances affecting people's opportunities
and treatment.

\section*{Limitations}
Reddit is not representative of all speakers or deployment settings.
Community provenance does not establish nationality, and topic, period,
moderation, and demographics can differ. Native-context comparisons
cannot isolate language or country effects. Binary gender labels and
White--Black and Christian--Muslim pairs omit substantial identity
diversity. Category ordering depends on the selected attributes and
retained subcorpora; no alternative-lexicon or topic-matched robustness
experiment is reported.

Absolute scores discard preference direction, and item-level confidence
intervals are not estimated. Rankings are descriptive, not tests of
population differences. Repetition dependence and conditioning on valid
responses further limit the independent binomial reference. Access-family
comparisons confound model scale, training, alignment, and serving.

\section*{Ethical Considerations}
Before annotation, MTurk workers received information about the study,
including a warning that excerpts could contain hostile or favourable
stereotypes and potentially distressing material. The statement described
voluntary participation, the option to decline answers or discontinue
without penalty, compensation through MTurk for completed tasks, and
academic-research use of the annotations. Consent was indicated by
proceeding after this information was presented. Appendix~\ref{app:consent}
reproduces the statement used before annotation. This study received
institutional ethics approval; identifying approval details are withheld
for anonymous review.

Annotator identifiers were removed and annotation results reported in
aggregate. Other identifiers were removed and verbatim text minimised. Public
visibility is not blanket permission to republish Reddit content, and
the paper's licence does not confer rights over third-party posts
\citep{reddit-developer-terms}. No unrestricted raw-text release is
provided. Any release of prompts, annotations, response counts, or code
requires explicit access terms, privacy safeguards, and applicable
retention and deletion procedures.

\bibliography{references}

\clearpage
\appendix
\section{MTurk Participant Information and Consent}
\label{app:consent}
The following statement was presented to MTurk workers before they
annotated the data reported in this paper. Its wording is reproduced here,
with formatting adapted to the paper layout.

\subsection*{Participant Information and Ethical Statement}
You are invited to participate in an academic research study examining
how large language models associate social groups with different
characteristics across languages and cultural contexts.

As part of this task, you will be asked to review short excerpts of
publicly available Reddit posts and classify whether they contain
stereotypes relating to gender, race, or religion. Some materials may
express negative or hostile stereotypes, while others may contain
seemingly favourable or positive stereotypes. These materials are
presented solely for research and evaluation purposes and do not represent
the views or beliefs of the researchers.

\noindent\textbf{Potential risks and discomfort:} Some excerpts may
contain sensitive, offensive, or potentially distressing content involving
social identities. Please be aware of this before proceeding. You may
decline to answer or discontinue participation if you feel uncomfortable.

\newpage
\noindent\textbf{Voluntary participation:} Your participation is entirely
voluntary. You are free to stop participating at any time, without penalty.

\noindent\textbf{Compensation:} You will receive compensation through
Amazon Mechanical Turk for completing the assigned annotation task.

\noindent\textbf{Privacy and confidentiality:} Your personal identity will
not be disclosed in any research publication. Annotator identifiers are
removed from the research dataset, and results are reported in aggregate.
The research focuses on annotation judgements rather than the personal
opinions, beliefs, or identities of individual participants.

\noindent\textbf{Data use:} The annotations collected will be used
exclusively for academic research on multilingual stereotype patterns and
the evaluation of large language models. The research findings may be
published in academic papers and related research outputs. Individual
annotators will not be identified in these publications.

By proceeding with this task, you acknowledge that you have read and
understood this information, are aware of the potentially sensitive nature
of the materials, and voluntarily agree to participate in the annotation
study.
\end{document}